\documentclass[letterpaper, 10 pt, conference]{ieeeconf}

\IEEEoverridecommandlockouts                              %

\usepackage[l2tabu,orthodox]{nag}

\usepackage[
    backend=biber,
    bibencoding=utf8,
    style=ieee,
    sorting=none,
    natbib=true,
    doi=false,
    isbn=false,
    url=false,
    eprint=false,
    maxcitenames=1,
    mincitenames=1,
]{biblatex}

\usepackage[draft,pdftex,colorlinks]{hyperref}
\usepackage[printonlyused]{acronym}

\usepackage{siunitx}
\usepackage[all]{nowidow}

\usepackage[dvipsnames]{xcolor}

\usepackage{lipsum}

\usepackage[pdftex]{graphicx}

\usepackage{epstopdf}

\usepackage{import}

\graphicspath{{./latexGoodPractices/}}

\usepackage{booktabs}

\usepackage{tabu}
\usepackage{makecell}
\usepackage{float}

\usepackage{amssymb,amsfonts,amsmath,amscd}

\usepackage{cleveref}
\creflabelformat{equation}{#2#1#3}

\usepackage{bm}

\newcommand{\bbm}{\begin{bmatrix}}
        \newcommand{\ebm}{\end{bmatrix}}

\usepackage[normalem]{ulem} %

\makeatletter
\usepackage[utf8]{inputenc}
\usepackage{commath}
\usepackage{mathtools}
\usepackage{mathabx}
\usepackage{siunitx}
\usepackage[ruled,vlined]{algorithm2e}
\usepackage[lofdepth,lotdepth]{subfig}
\usepackage{tikz}
\usepackage{microtype}

\let\oldtheequation\theequation
\renewcommand\tagform@[1]{\maketag@@@{\ignorespaces#1\unskip\@@italiccorr}}
\renewcommand\theequation{(\oldtheequation)}
\makeatother

\usepackage[belowskip=0pt,aboveskip=1pt]{caption}
\graphicspath{{media/}}

\newcommand{\C}{\bm{C}}

\newcommand{\J}{\bm{J}}
\newcommand{\E}{\mathbb{E}}
\newcommand{\I}{\bm{1}}

\newcommand{\R}{\mathbb{R}}
\newcommand{\N}{\mathbb{N}}

\usepackage{amssymb}%
\usepackage{pifont}%

\acrodef{ICP}{Iterative Closest Point}
\acrodef{DOF}{Degrees Of Freedom}
\acrodef{3-DOF}{Three Degrees Of Freedom}
\acrodef{4-DOF}{Four Degrees Of Freedom}
\acrodef{6-DOF}{Six Degrees Of Freedom}
\acrodef{INS}{Inertial Navigation System}
\acrodef{GNSS}{Global Navigation Satellite System}
\acrodef{GPS}{Global Positioning System}
\acrodef{UGV}{Unmanned Ground Vehicle}
\acrodef{UAV}{Unmanned Aerial Vehicle}
\acrodef{MAV}{Micro Aerial Vehicle}
\acrodef{IMU}{Inertial Measurement Unit}
\acrodef{SLAM}{Simultaneous Localization and Mapping}
\acrodef{MEMS}{Micro-Electromechanical Systems}

\DeclareUnicodeCharacter{0308}{¨} %

\usepackage{fancyhdr}
\fancypagestyle{withfooter}{
  
  \fancyfoot[C]{\footnotesize Accepted to the IEEE ICRA Workshop of Field Robotics 2024}
}

\begin{document}
\title{\LARGE \textbf{3D Mapping of Glacier Moulins: Challenges and lessons learned}}

\author{William Dubois,$^{1}$ Matěj Boxan,$^1$ Johann Laconte,$^2$ and François Pomerleau$^{1}$%
  \thanks{$^{1}$Northern Robotics Laboratory, Université Laval, Québec City, Canada,
    {\texttt{\small{$\{$william.dubois, matej.boxan, francois.pomerleau$\}$ @norlab.ulaval.ca}}}}%
  \thanks{Université Clermont Auvergne, INRAE, UR TSCF, 63000, Clermont-
Ferrand, France {\texttt{\small{johann.laconte@inrae.fr}}}}%
}

\linepenalty=3000
\addtolength{\textfloatsep}{-0.1in}

\maketitle
\thispagestyle{withfooter}
\pagestyle{withfooter}

\begin{abstract}
In this paper, we present a field report of the mapping of the Athabasca Glacier, using a custom-made lidar-inertial mapping platform. 
With the increasing autonomy of robotics, a wider spectrum of applications emerges.
Among these, the surveying of environmental areas presents arduous and hazardous challenges for human operators. 
Leveraging automated platforms for data collection holds the promise of unlocking new applications and a deeper comprehension of the environment. 
Over the course of a week-long deployment, we collected glacier data using a tailor-made measurement platform and reflected on the inherent challenges associated with such experiments.
We focus on the insights gained and the forthcoming challenges that robotics must surmount to effectively map these terrains.
\end{abstract}

\acresetall
\section{Introduction}

Deploying robots in the cryosphere is still an open problem, particularly when it comes to localization and mapping~\cite{Pomerleau2023}.
These deployments serve as crucial sources of data collection and analysis, essential for understanding climate change.
However, accessing certain environments poses significant challenges and risks to human safety. 
In this context, robotics emerges as a viable solution, enabling the deployment of automated platforms to navigate hazardous settings and gather environmental data. 
Notably, Dante was the first robot successfully deployed in a remote Alaskan volcano~\cite{Dante2}, demonstrating the feasibility of this approach.
Similarly, glacier monitoring stands as a critical endeavor due to the adverse impacts of melting ice. 

Yet, conducting surveys within glaciers remains hazardous and logistically arduous. 
Robotic systems have the growing potential to undertake such tasks.
For example, \citet{Talbot2023} describe a robotic deployment in degraded glacier settings, focusing on the lidar-based mapping approach validated on the Athabasca Glacier, Canada.
In parallel, \citet{Polzin2023} further show the utility of robotics in extreme environments, particularly in glacial moulin exploration in the \textit{Mer de Glace} glacier at Mont Blanc.
A moulin is a near-vertical cave carved by melting water streaming down a glacier.
The study emphasizes the risks associated with human-led data gathering and highlights the promising role of robotic platforms in monitoring changes in glacial environments. 
Because of the inherent challenges tied to these extreme environments, such as the lack of distinct features, and the oscillation of the tethered robot, further research is necessary before achieving true autonomy.

As such, we present in this short paper a deployment of a robotic platform in the Athabasca Glacier, focusing on the forthcoming research opportunities and lessons learned.

\begin{figure}[t]
\centering
\includegraphics[width=\linewidth]{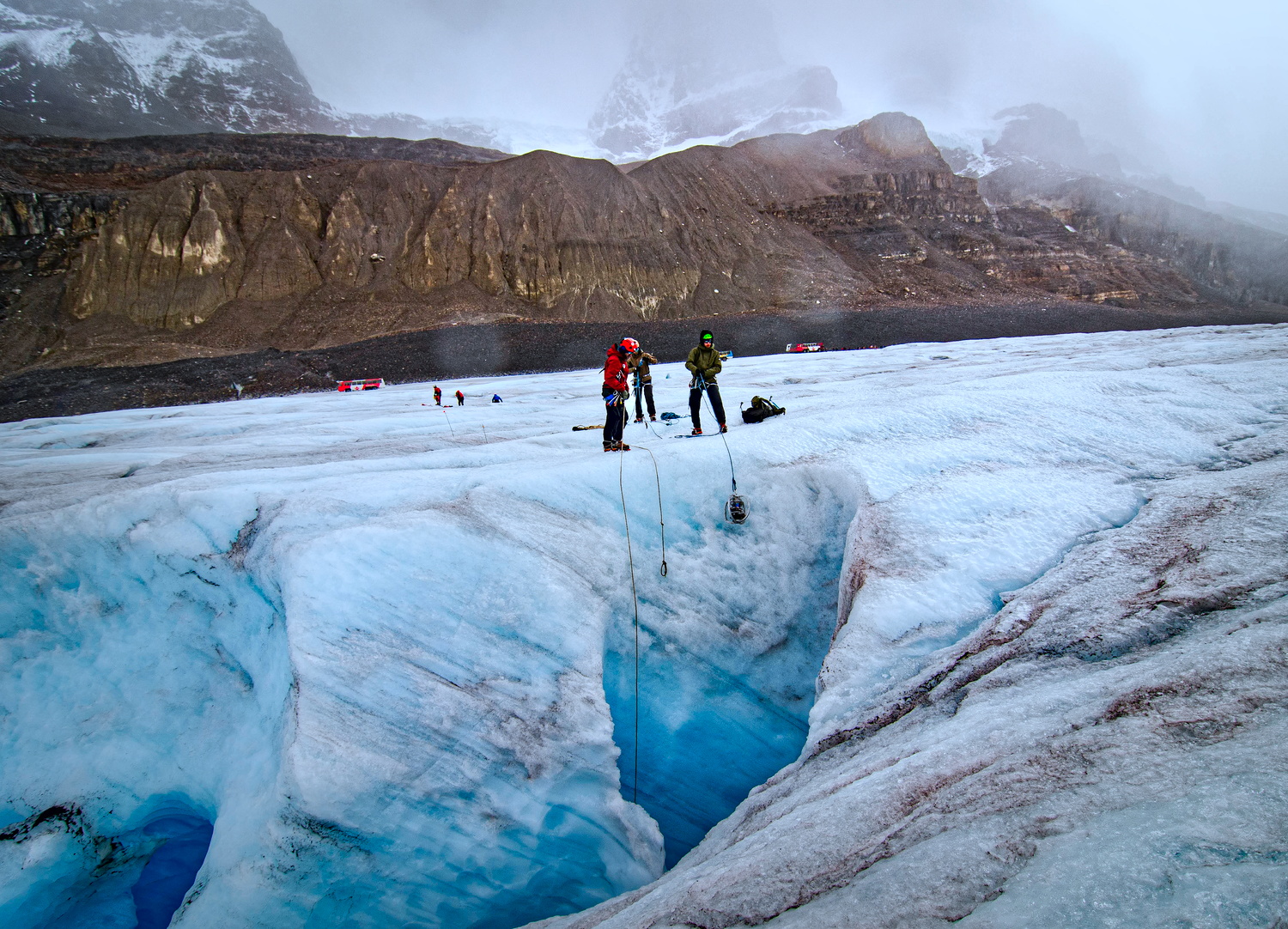}
\caption{Deployment conducted on the Athabasca Glacier. The experimental platform was lowered in a glacial moulin, mapping its surroundings.}~\label{fig:fig-intro}
\end{figure}

\section{Experiments}
The deployment took place on the Athabasca Glacier, located within the rugged landscape of the Canadian Rockies. 
This glacier is characterized by its complex features such as ice canyons, crevasses, and large glacial moulins, making it a high-risk environment. 
Safety measures are crucial, requiring careful anchoring systems to safely approach the edges of the glacier's moulins, which serve as entry points to its inner layers. 
The process of descending into these icy chasms necessitates thorough preparation, including the time-consuming task of establishing and securing lifelines and anchors, ensuring the safety of personnel and equipment amidst the challenging surroundings.
In order to alleviate the risks, a data-gathering platform was tethered inside the moulins, as shown in \autoref{fig:fig-intro}.

\begin{figure}[thbp]
\centering
\includegraphics[width=\linewidth]{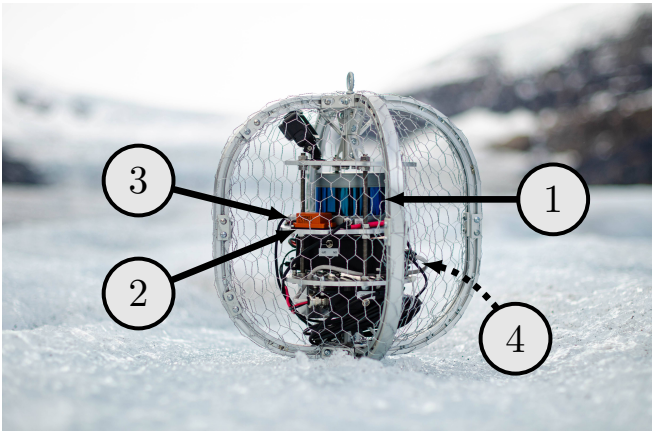}
\caption{Data gathering platform with which sensor measurements were recorded to perform 3D localization and mapping, equipped with a lidar Robosense RS-16 (\texttt{1}), an Xsens MTi-10 \ac{IMU} (\texttt{2}), a Vectornav vn100 \ac{IMU} (\texttt{3}, behind the Xsens MTi-10) and a barometric pressure sensor DPS310 (\texttt{4}, on the other side of the platform)}
\label{fig:fig-experimental-setup}
\end{figure}

\begin{figure}[t]
\centering
\includegraphics[width=\linewidth, clip, trim={0 0 0 0}]{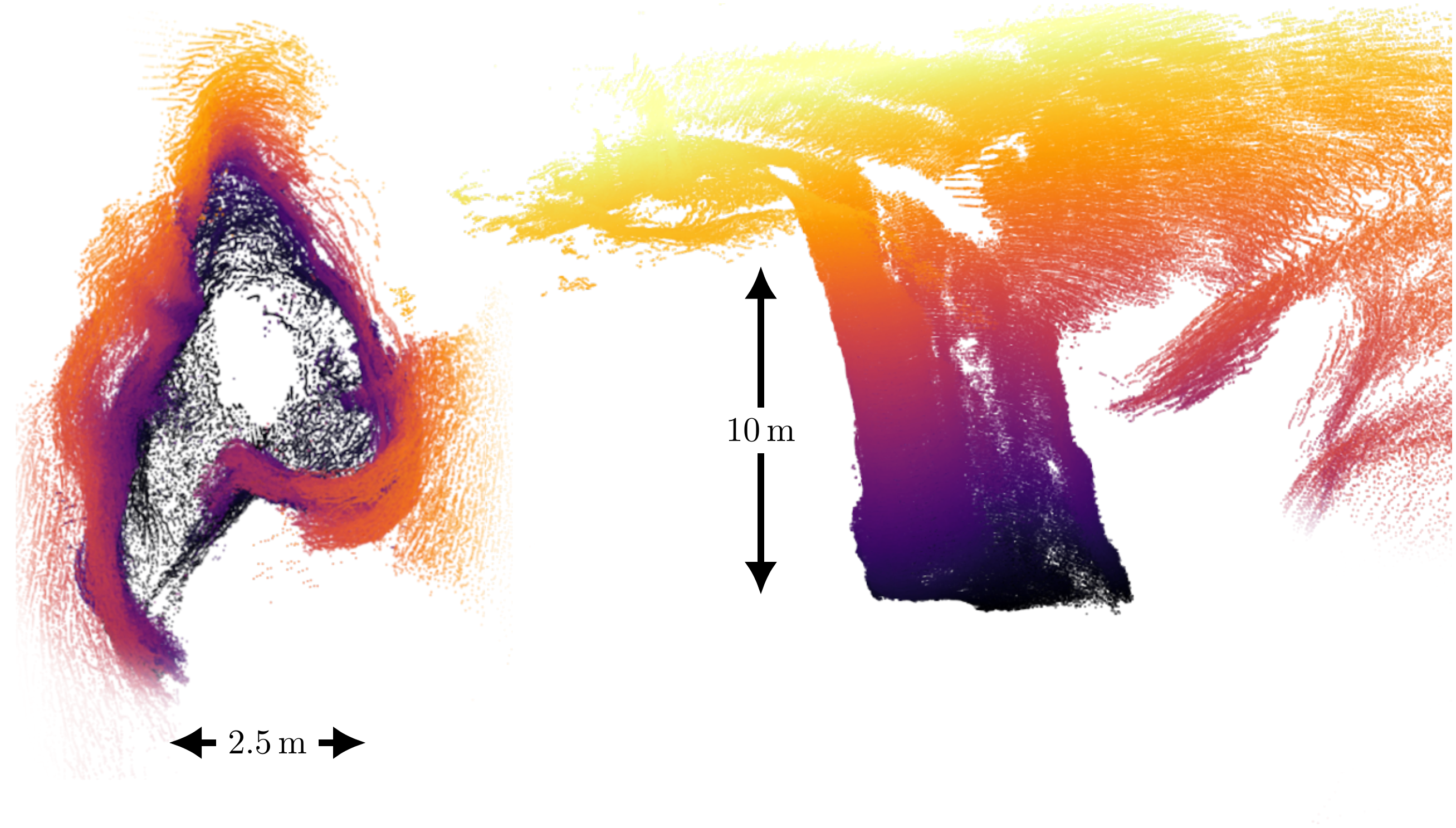}
\caption{Map result (colored by elevation) from the moulin experiment. Left: top view; Right: Side view. The lack of features in the moulin makes the mapping of such environments challenging.}~\label{fig:map-results}
\end{figure}

The platform used to record data is depicted in \autoref{fig:fig-experimental-setup}.
It is equipped with a lidar (Robosense RS-16), two \acp{IMU} (Xsens MTi-10, Vectornav vn100), and a barometric pressure sensor (DPS310).
The platform was manually lowered inside the glacier's moulin, with the operators staying at least at arm's length from the opening.
A 3D map of the environment is generated using the lidar data, with an example shown in \autoref{fig:map-results}.
Along with the adverse conditions of the environment, the moulin displays little features for \ac{ICP} to converge.
In the next section, we discuss the main challenges that remain to map such an environment, as well as the practical issues linked to deployments in arctic regions.

\section{Challenges and Lessons Learned}
While recording data in the extreme environment of the Athabasca Glacier, we encountered numerous obstacles, necessitating robust solutions and meticulous adherence to safety protocols. 
Weather conditions were very erratic.
Within a few hours, operators sustained sudden shifts ranging from hail, rain, and fog to snow and freezing rain, highlighting the unpredictable nature of the environment. 
To withstand such conditions, rugged laptops are indispensable; however, a setback arose when our designated laptop malfunctioned prior to deployment, compelling us to switch to an office laptop on the glacier.
Even with resilient, rugged equipment, it is thus necessary to always prepare a spare set of equipment, which becomes more and more critical as the deployment takes place in remote areas.

Furthermore, deploying rugged platforms emerged as imperative, with our experience revealing a vulnerability in the form of a non-rugged computer (Raspberry Pi 4B) on the platform. 
Although, as a whole, the platform exhibited strong mechanical robustness, the connector for the SD card reader weakened after multiple high-velocity shocks, leading to the system not booting. 
A solution is to investigate embedded computers following the PC/104 standard.
On one side, these computers were used successfully in micro-satellites, but on the other, they are very specialized and hard to find with a generic operating system such as Ubuntu.
As the environment showcases hazardous places, adherence to stringent safety measures remained non-negotiable, with no exceptions tolerated.
While such measures are necessary, this clearly highlights the need to prepare the deployment in advance, with every subsystem and procedure meticulously tested beforehand. 

Additionally, the environmental impact of the deployment has to be taken into account, ensuring no pollution will be left on-site, and no deterioration of the environment will come from the field deployment. 
In our case, a metal cage around the platform served as a precautionary measure in case of material loss due to mechanical failure, wear, and tears.

The difficulties encountered also affected mapping solutions, as the lack of distinct features in the environment led to under-constrained and degraded solutions. 
Furthermore, as the platform was lowered in the moulin, sudden movements and jerks of the platform, caused by collisions and inconsistent lowering speeds, hindered the mapping capabilities, making it challenging to map in real-time and thus impeding the on-site monitoring of the system's well-being.
Ultimately, these conditions led to the necessity to compute the map offline at a much lower rate, since the lack of features, and therefore constraints, and the high-velocity movements prevented the \ac{ICP} registration from converging.

From this deployment, several key lessons emerged. 
Foremost among these was the paramount importance of meticulous preparation, underscored by the necessity of comprehensive testing and validation of equipment and procedures prior to deployment. 
A meticulously documented experimental procedure is essential, given the abundance of information to assimilate during deployment, mitigating the risk of errors. 
Furthermore, a robust verification procedure is paramount, as frequent on-site validation of the data will mitigate losses in case of a system failure.

\section{Conclusion}
In this short paper, we describe the deployment of a mapping platform in the Athabasca Glacier.
The harsh conditions of arctic environments make deployments arduous, with their specific set of challenges.
Preparation, resilience, and on-site monitoring of the system are key for the success of the deployment.
The specific environment of a glacier moulin led to interesting challenges, which are the reflectiveness of the ice coupled with a clear lack of features.
Future work will focus on designing a new version of the measurement platform and fusing information from several sensors, such as atmospheric pressure, to improve the robustness of the mapping process.

\section*{Acknowledgment}

This research was supported by the  Natural  Sciences and Engineering  Research  Council of  Canada  (NSERC)  through the General Research Fund (GRF) from Université Laval.

\renewcommand*{\bibfont}{\small}
\IEEEtriggeratref{6}
\IEEEtriggercmd{\enlargethispage{-0.1in}}

\printbibliography

\end{document}